\pdfoutput=1

\documentclass[11pt]{article}

\usepackage[]{EMNLP2022}

\usepackage{times}
\usepackage{latexsym}

\usepackage[T1]{fontenc}

\usepackage[utf8]{inputenc}

\usepackage{microtype}

\usepackage{inconsolata}

\usepackage{multirow}
\usepackage{graphicx}
\usepackage{subcaption}
\usepackage{amssymb}
\usepackage{amsmath}
\usepackage{bbm}
\usepackage{enumitem}
\usepackage{float} 
\usepackage[flushleft]{threeparttable}
\usepackage{arydshln}
\title{TranSHER: Translating Knowledge Graph Embedding with Hyper-Ellipsoidal Restriction}

\newcommand{\comment}[1]{}

\author{ 
    Yizhi Li\textsuperscript{1}\thanks{\quad Part of the work is done at Pingan Technology.}\space, 
    Wei Fan\textsuperscript{2}\space, 
    Chao Liu\textsuperscript{3},
    Chenghua Lin\textsuperscript{1}\thanks{\quad Corresponding author.}\;,
    Jiang Qian\textsuperscript{3} \\ 
    {
    \textsuperscript{1} Department of Computer Science, The University of Sheffield, UK
    }\\ 
    {
    \textsuperscript{2} Department of Computer Science, University of Central Florida, USA
    }\\
    {
    \textsuperscript{3} Pingan Technology, China
    }\\ 
    \texttt{
        \{yizhi.li, c.lin\}@sheffield.ac.uk, weifan@knights.ucf.edu, 
    } \\ 
    \texttt{
        lliuchao666@mail.ustc.edu.cn, jqian104@126.com
    }
}

\begin{document}
\maketitle

\begin{abstract}

Knowledge graph embedding methods are important for the knowledge graph completion (or link prediction) task.
One existing efficient method, PairRE, leverages two separate vectors to model complex relations (i.e., 1-to-N, N-to-1, and N-to-N) in knowledge graphs.
However, such a method strictly restricts entities on the hyper-ellipsoid surfaces which limits the optimization of entity distribution, leading to suboptimal performance of knowledge graph completion.
To address this issue, we propose a novel score function \textit{TranSHER}, which leverages relation-specific translations between head and tail entities to relax the constraint of hyper-ellipsoid restrictions. 
By introducing an intuitive and simple relation-specific translation, \textit{TranSHER} can provide more direct guidance on optimization and capture more semantic characteristics of entities with complex relations. 
Experimental results show that \textit{TranSHER} achieves significant performance improvements on link prediction and generalizes well to datasets in different domains and scales. 
Our codes are public available at \url{https://github.com/yizhilll/TranSHER}.
\end{abstract}

\vspace{-1mm}
\section{Introduction}

Knowledge graph is proposed to structurally store human knowledge when the development of computer science has brought exponentially growing digitized information. Knowledge graphs have been widely adopted in important applications, such as semantic parsing~\cite{berant2013semantic_parsing_kg}, question generation and answering~\cite{lin2015sherlock,hao2017end_qa_kg,peng2021named}, and information retrieval~\cite{xiong2017explicit_ir_kg}. However, due to the expanding nature and difficulties of construction, knowledge graphs often suffer from incompleteness. Thus, knowledge graph completion (or link prediction) becomes a task of great importance from both a research and application perspective.

\begin{figure}[tb]
\centering

{\centering
\begin{subfigure}{0.46\linewidth}
\centering
\includegraphics[width=\linewidth]{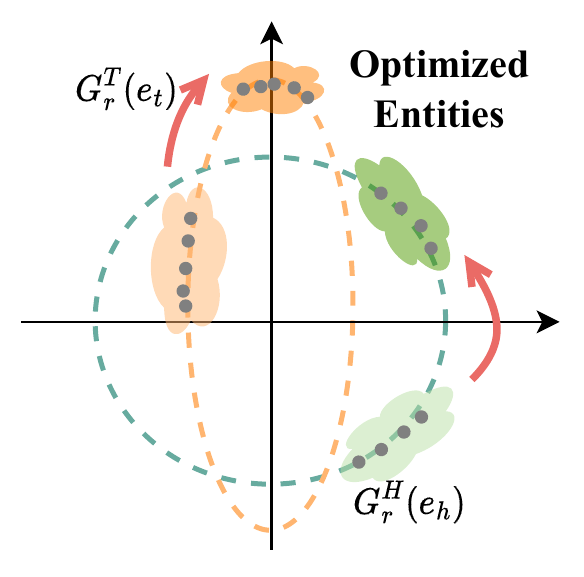} 
\caption{PairRE}\label{fig_insight_a}
\end{subfigure}
\begin{subfigure}{0.46\linewidth}
\centering
\includegraphics[width=\linewidth]{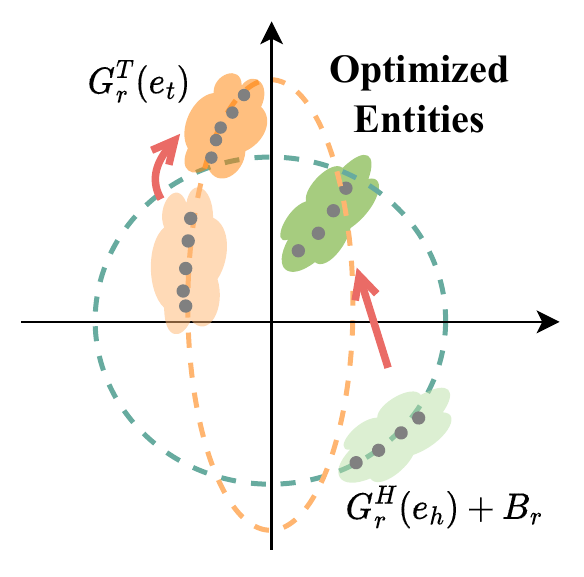} 
\caption{TranSHER}\label{fig_insight_b}
\end{subfigure}
}
\caption{
A Comparison Example of Distribution Optimization Between PairRE and TranSHER in 2-dimensional Space. 
}

\label{fig_insight}
\end{figure}

To construct a knowledge graph, entities and relations extracted from facts are treated as nodes and edges.
A single fact in this graph is represented as a directed relation-specific link between two entities, which is denoted as a triplet like \textit{(head, relation, tail)}. 
Since knowledge graphs usually contain a large number of entities and relations, knowledge graph embeddings are proposed to efficiently learn the representation of graphs and accomplish the link prediction task with \textit{score functions}. Intuitions behind the design of score functions can be summarized as two important principles: 1) logical reasoning \textit{relation patterns} such as symmetry (antisymmetry), inversion, and composition; 2) statistically categorized \textit{relation types} like 1-to-1, 1-to-N, N-to-1, and N-to-N, where the latter three are called \textit{complex relations}.

Inspired by word2vec~\cite{mikolov2013distributed}, TransE~\cite{bordes_translating_2013_TransE_FB15k} represents knowledge graphs with relation-specific translations between head and tail entities to model the connections, where the score function works upon the translational distance $\|e_h+r-e_t\|$.
Following this design, certain works~\cite{wang2014TransH, lin2015learning_TransR, Xiao2016_ManifoldE} improve TransE by conducting relation-specific transformations on the entities before calculating distance. 
Certain works~\cite{wang2014TransH, trouillon2016complex, sun2018rotate, chao_pairre_2021} further apply the principles of modeling relation patterns and complex relations to refine the design of score functions.

Among these methods, the recently proposed PairRE ~\cite{chao_pairre_2021} uses separate relation vectors for head and tail entities to better model the complex relations and outperforms preceding methods. 
However, after our preliminary exploration, we notice that PairRE imposes a strong restriction that fixes the $L_2$ norm of the  entity vector $\vec{e}$, and scales it with coefficients along all dimensions. 
Under such a restriction, the entities are essentially distributed on the surface of hyper-ellipsoids, where the foci are exactly laying on the axes. 
For the same reason, entity embeddings are strictly limited and can only be optimized around the hyper-ellipsoidal surface. %
In other words, entity distribution can only move along an arc path to match the true connections, as Fig. \ref{fig_insight_a} shows, which may impose the close entity embeddings entangled and bring difficulties to the modeling.

To tackle the aforementioned challenges, we propose \textit{TranSHER} which has a novel score function that leverages relation-specific translations between head and tail entities to relax the constraint of hyper-ellipsoid restrictions. 
We hypothesize that introducing translations for entities of PairRE can simplify the multi-relation modeling of entities on hyper-ellipsoids by providing an extra degree of freedom and thus improve the distribution learning of entities. 
In order to model complex relations, TranSHER first follows PairRE to conduct mappings on head and tail entities separately.
Then, TranSHER additionally performs relation-specific translations on entities while holding the hyper-ellipsoidal restriction. 
Take Fig. \ref{fig_insight_b} as an example: the mapped head entities (green) are moved closer to the tail entities (orange) without requiring obvious changes of the entities within the cluster and form a better distribution for relational modeling because the translational distance provides more direct guidance in the score function.
The relation-specific translation in our score function allows more flexible optimization by relaxing the arc path constraint in PairRE brought by the hyper-ellipsoidal mapping, leading to a better distribution of entities and knowledge graph completion.

Moreover, TranSHER can better model the semantic characteristics of entities and utilize them to improve entity retrieval for complex relations.
The semantic characteristics usually decide the categories of entities (e.g., people or companies); TranSHER can model the connected structure between entities of different categories more accurately in complex relations.
For those complex relation triplets, TranSHER further enhances link prediction by retrieving multiple candidate entities in the correct category. 
For example, for a tail prediction query (film, produced\_by, ?), TranSHER ranks the names of film producers at the front while other models may be confused by the related entertainment company or studio (cf. \S\ref{sec_case}). %

In addition, while achieving qualitative embeddings, TranSHER can further maintain the ability to model important relation patterns, namely, symmetry (antisymmetry), inversion, and composition under certain constraints (cf. \S\ref{sec_method}). With such ability, TranSHER has the potential to be generalized well on datasets from different knowledge domains in different real-world settings. 

We conduct comprehensive experiments across five datasets from different domains, which demonstrate the effectiveness of TranSHER. Impressively, TranSHER has achieved substantial improvement compared with the best baseline:
the MRR increase has reached up to 4.6\% on YAGO37 and 3.2\% on ogbl-wikikg2.
We also conduct many analytical experiments to study how translations of TranSHER behave and enhance knowledge graph completion. Some case studies further demonstrate the superiority of our approach.

\section{Related Work}

Knowledge graph embedding methods are proposed to model the intrinsic properties of facts and to conduct knowledge graph completion. To further complete knowledge graphs, these methods use designed score functions to model complex relations and various patterns. By measuring the relation-related distance between entities, KGE models predict the probabilities of given triplet queries $(e_h,r,?)$ or $(?,r,e_t)$ to complete the graph. 
Such models are characterized by their corresponding score functions and the embeddings are usually optimized with gradient descent algorithms.

\noindent\textbf{Distance-based} score functions model the triplet facts by calculating distances between entity embeddings in the Euclidean space. Proposed by TransE~\cite{bordes_translating_2013_TransE_FB15k}, one popular practice is to conduct a relation-specific translation on the given entity before distance calculation, i.e., let $e_h+r\approx e_t$ in the case where the fact holds.
Such a translational principle empowers the models to conduct knowledge graph completion on large-scale datasets while maintaining their performance. Many score functions, such as TransH, TransR, and ManifoldE~\cite{wang2014TransH, lin2015learning_TransR, Xiao2016_ManifoldE}, followed this principle of translation and achieved fair performances. However, RotatE~\cite{sun2018rotate} claims these extended works of TransE are weak in modeling certain relation patterns and thus proposes a solution of modeling in the complex space. 
Moreover, PairRE~\cite{chao_pairre_2021} argues that complex relations can be better modeled by separating relation vectors for heads and tails. 
Our TranSHER aims to provide a more effective model for complex relations by bridging the gap between translational distance-based models and the latest model PairRE.

%
\noindent\textbf{Semantic matching} score functions aim to predict the existence of facts by measuring the semantic similarity among entities and the given relation in the same representation space. RESCAL~\cite{nickel2011three_rescal} introduces a bilinear function $h^\top M_r t$ to represent the similarity score but suffers from modeling complexity. Some following work such as DistMult, HolE, and ComplEx~\cite{yang2014embedding_distmult, nickel2016holographic_hole, trouillon2016complex} intend to simplify the model while preserving critical features. A more recent work SEEK~\cite{xu-etal-2020-seek} generalizes the existing semantic matching score functions by segmenting the embedding to facilitate feature interactions, while keeping the same model size. In general, the semantic matching models struggled to distinguish similar entities and lack the ability to simultaneously model multiple relation patterns. Our TranSHER model tackles these challenges by proposing a novel score function which leverages relation-specific translations, yielding an effective improvement in modeling complex relations.

\section{Problem Formulation}

In this section, we describe the task definition and notations for better illustration. 
The set of known facts in a knowledge graph is represented by $\mathcal{T}$, which includes triplets $(e_h,r,e_t)$.
The notation $(e_h,r,e_t) \in \mathcal{T}$ will be used when the fact holds. The set of entity $e$ and the set of relation $r$ are denoted as $\mathcal{E}$ and $\mathcal{R}$. 

Knowledge graph completion (also regarded as link prediction) aims to predict the missing links of knowledge graphs. Specifically, given a triplet $(e_h,r,e_{t'})$ or $(e_{h'},r,e_t)$, a score function $f_r$ is required to output an existence probability of the triplet.
Since all the entities in $\mathcal{E}$ are provided in the training set, knowledge graph completion can also be regarded as a ranking task. For the true candidate entities, models are expected to assign higher rankings than false ones.

For a given entity-relation query $(e_h,r,?)$ or $(?,r,e_t)$, there may exist multiple correct answers to complete the triplet, i.e., the quantities of those entities satisfy $\|\{e_{t'}|(h,r,e_{t'})\in \mathcal{T} \}\|>1$ or $\|\{e_{h'}|(e_{h'},r,e_t)\in \mathcal{T} \}\|>1$. 
According to the average \textit{heads per tail} and \textit{tails per head} counted through the dataset, relations in $\mathcal{R}$ can be categorized into 4 types: 1-to-1, 1-to-N, N-to-1, and N-to-N~\cite{wang2014TransH}. 

Relations can also be summarized by several significant patterns: symmetry/antisymmetry, inversion, and composition. The definitions are given as follows:
\begin{itemize}[itemsep=0pt,topsep=3pt]
    \item A relation $r$ is \textbf{symmetric} or \textbf{antisymmetric} if $ (e_1,r,e_2)\in \mathcal{T}, \forall e_1,e_2 \in \mathcal{E} \Rightarrow (e_2,r,e_1) \in \mathcal{T}$ or $(e_2,r,e_1) \not\in \mathcal{T}$. 
    \item Relation $r_1$ and relation $r_2$ are \textbf{inverse} if $ (e_1,r_1,e_2)\in \mathcal{T}, \forall e_1,e_2 \in \mathcal{E} \Rightarrow (e_2,r_2,e_1) \in \mathcal{T}$. 
    \item Relation $r_3$ is \textbf{composed} by relation $r_1$ and $r_2$ if $ (e_1,r_1,e_2)\in \mathcal{T} \land  (e_2,r_2,e_3)\in \mathcal{T} , \forall e_1,e_2,e_3 \in \mathcal{E} \Rightarrow (e_1,r_3,e_3) \in \mathcal{T}$.
\end{itemize}

\section{TranSHER}
\label{sec_method}

\subsection{Score Function}

We propose a simple yet effective translational distance-based score function \textit{TranSHER}. The key intuition behind TranSHER is to provide more freedom with the relation-specific translation to ease the hyper-ellipsoidal restriction, while still keeping enough ability for complex relations modeling and training stability.

For this aim, TranSHER first maps the entity vectors to hyper-ellipsoids with underlying fixed-norm restriction for the actual entity embeddings and hence brings about general training stability; then conducts a relation-specific translation on the restricted entities for modeling the distances between mapped head and tail clusters.
Specifically, we first define a mapping function $G(e)$ to restrict the entities on the hyper-ellipsoid surface. Since the fact triplets are directional, we use two separate relation-specific mapping functions $G_r^H (e_h)$ and $G_r^T (e_t)$ to manage the cases when an entity is considered as head or tail, correspondingly:
\begin{align}
    & G_r^H (e_h) = r^H \circ \frac{e_h}{|e_h|}, r^H, h \in \mathbb{R}^k  \\ 
    & G_r^T (e_t) = r^T \circ \frac{e_t}{|e_t|}, r^T, t \in \mathbb{R}^k  
\end{align}
where the $\circ$ here stands for the element-wise product. The mapping can be regarded as restricting the entities on unit hyper-spheres by fixing the $L_2$ norm $||\vec{e}||_2 = 1$ and conducting further linear scaling w.r.t. different relations. As the softmax loss in training intends to learn radially distributed entity representations~\cite{Wang2017NormFaceL2}, such a fixed-norm restriction of entity representations consequently benefits the stability of TranSHER optimization process~\cite{xu2018spherical, wang2020understanding}.
With $G_r^H$ and $G_r^T$, we are able to map the entity vectors to two hyper-ellipsoids according to the relations and whether they are heads or tails, as shown in Fig. \ref{fig_insight_a}. 
Note that the entity embeddings will distribute on the unit hyper-sphere if $r^H = \mathbbm{1}$ or $r^T = \mathbbm{1}$.


Then, we introduce a relation-specific translation item $B_r \in \mathbb{R}^k$, which not only eases the hard hyper-ellipsoidal restriction but also encourages to identify the hard-to-distinguish entities close in space for complex relations. 
Accordingly, the final score function of TranSHER can be derived as:
\begin{equation}
   f_r(e_h,e_t) = \gamma - || G_{r}^{H}(e_h) + B_r - G_{r}^{T}(e_t)||_{1}
\label{eq_core_sf}
\end{equation}
where $\gamma$ is an adjustable constant margin. Note that all the embeddings share the same dimension setting $k$, i.e., $r^H, r^T, B_r, e \in \mathbb{R}^k$.
The additional translation for the entities with hyper-ellipsoidal restriction increases the degree of freedom of the score function and hence could provide extra optimization options for modeling the distances between entity clusters connected by complex relations.

\subsection{Initialization}\label{sec_method_init}
The initialization strategy plays an important role in neural network optimization~\cite{Erhan09Difficulty, hayou2018selection}, especially in the case of knowledge graph modeling that consists of enormous numbers of entity-relation-entity interactions. Classic knowledge graph embedding works pay less attention to initialization and adopt simple strategies. For example, both TransE and PairRE randomly initialize all the embedding weights with the same uniform distribution. After our practical implementation, we notice different initialization strategies place different assumptions on embeddings, which also largely influence the performances. 

In this regard, all three main components in TranSHER need to be well-considered and initialized, i.e., the relation embeddings $\mathcal{R}$, entity embeddings $\mathcal{E}$, and translations $B_r$. 
To better model the distribution of knowledge graph embeddings, we propose to conduct initialization searching for the optimal initial distributions for each component in TranSHER.
We select empirically validated strategies from two main categories, uniform, and normal distributions, for the component-independent initialization of TranSHER.
For uniform distribution we follow the setting in~\citet{sun2018rotate}, where the weights are initialized with the gamma uniform $ \mathcal{U}_{\gamma}( -\frac{\gamma+\epsilon}{k}, \frac{\gamma+\epsilon}{k})$. The $\epsilon$ here is a placeholder constant for the edge case that $\gamma=0$. 
For the normal distribution, the Xavier normal proposed in~\citet{glorot2010understanding_xavier} is adapted as $\mathcal{N}_{X}(0,g \cdot \sqrt{\frac{2}{k}})$ for TranSHER, where the gain $g$ is defined as a scaling factor.

Without strict distribution assumptions for the parameters in the three components (i.e., $\mathcal{R}$, ${E}$, and $B_r$), TranSHER allows the components to be initialized independently with different distributions.
The selection of the initial distribution for each component in TranSHER can vary through different knowledge graphs since they do not share the same data distribution. We will discuss in section~\ref{sec_result_bias_init} that, due to a more effective optimization process and potentially introducing appropriate inductive bias, such an initialization searching strategy allows TranSHER to produce better results.

\subsection{Training and Optimization}
Following the standard self-adversarial training framework~\cite{sun2018rotate}, the general loss function for optimization is:
\begin{align}
    \mathcal{L} = 
    & -\log\sigma(f_r(e_h,e_t))  \notag\\
    & + \sum_{i}^{N} p(e_{h_i'},r,e_{t_i'}) \log\sigma(-f_{r}(e_{h_i'},e_{t_i'}))
\end{align}
where $\sigma$ stands for the sigmoid activate function and $p(e_{h_i'},r,e_{t_i'})$ is the self-adversarial weight calculated according to the scores.

\subsection{Modeling Ability}
Given the aforementioned design, our score function can be better optimized in training without losing the ability to learn different relation patterns. We prove that TranSHER can model symmetric/antisymmetric, inverse, and composed relations with the following constraints:
\begin{itemize}[itemsep=1pt,topsep=5pt] 
    \item \textbf{symmetry}:
        $(r_1^H\circ \frac{e_1}{|e_1|} + B_{r_1} - r_1^T \circ \frac{e_2}{|e_2|}=0)\land(r_1^H\circ \frac{e_2}{|e_2|} + B_{r_1} - r_1^T \circ \frac{e_1}{|e_1|}=0)
        \Rightarrow$ $r_1^T=-r_1^H$
    \item \textbf{antisymmetry}: 
        $(r_1^H\circ \frac{e_1}{|e_1|} + B_{r_1} - r_1^T \circ \frac{e_2}{|e_2|}=0)\land(r_1^H\circ \frac{e_2}{|e_2|} + B_{r_1} - r_1^T \circ \frac{e_1}{|e_1|}\not=0)
        \Rightarrow r_1^T\not=-r_1^H$
    \item \textbf{inversion}:
        $(r_1^H\circ \frac{e_1}{|e_1|} + B_{r_1} - r_1^T \circ \frac{e_2}{|e_2|}=0) \land (r_2^H\circ \frac{e_2}{|e_2|} + B_{r_2} - r_2^T \circ \frac{e_1}{|e_1|}=0)
        \Rightarrow (r_1^T\circ r_2^T = r_1^H \circ r_2^H) \land (B_{r_1} = - B_{r_2} ) $
    \item \textbf{composition}:
        $(r_1^H\circ \frac{e_1}{|e_1|} + B_{r_1} - r_1^T \circ \frac{e_2}{|e_2|}=0)\land 
        (r_2^H\circ \frac{e_2}{|e_2|} + B_{r_2} - r_2^T \circ \frac{e_3}{|e_3|}=0) \land
        (r_3^H\circ \frac{e_1}{|e_1|} + B_{r_3} - r_3^T \circ \frac{e_3}{|e_3|}=0)
        \Rightarrow (r_3^H=r_1^H\circ r_2^H) \land (r_3^T=r_1^T\circ r_2^T) \land (B_{r_3}=r_2^H\circ B_{r_1}+r_1^T\circ B_{r_2})$
\end{itemize}

This proof shows that TranSHER introduces additional relation-specific translations for easing the hyper-ellipsoidal restriction without losing the ability of modeling various relation patterns.

\section{Experiments}
\subsection{Experimental Setup}
\noindent \textbf{Datasets and Evaluation.} 
We conduct extensive experimentation on five publicly available datasets for two evaluation settings of the link prediction task. 
For the classic full ranking setting that requires an exhaustive search through the entity set $\mathcal{E}$, we select FB15k-237~\cite{toutanova_observed_2015_fb15k237}, YAGO37~\cite{guo2018knowledge_RUGE_YAGO37}, and DB100K~\cite{ding-etal-2018-improving-db100k}, which are extracted and constructed from knowledge databases~\cite{suchanek2007yago,bollacker2008freebase, bizer2009dbpedia}.
Adopted from the Open Graph Benchmark (OGB)~\cite{hu2020_OGB}, the other setting of partial ranking only requires distinguishing the target entity from a randomly sampled entity subset. The ogbl-wikikg2 dataset with a massive number of triplets and the ogbl-biokg dataset consisting of biomedical facts are selected from the OGB link property prediction leaderboard. 
The statistics of the dataset are listed in Tab.~\ref{tbl_dataset}. 
For both evaluation settings, the Mean Reciprocal Rank (MRR) is regarded as the main metric and Hits at N (HIT@N) as the auxiliary metric.
More details of the datasets and evaluation protocol can be referred to in Appendix~\ref{sec:appendix_a_data} and~\ref{sec:appendix_a_eval}.
\begin{table}[tb]

\scalebox{0.83}{
    \begin{tabular}{l|ccccc}
    \hline
    \textbf{Dataset} & \textbf{Rel.} & \textbf{Ent.} & \textbf{Train} & \textbf{Valid} & \textbf{Test}\\
    \hline 
    FB15k-237 & 237 & 15k & 272k & 18k & 20k \\
    DB100K & 470 & 100k & 598k & 50k & 50k \\
    YAGO37 & 37 & 123k & 989k & 50k & 50k \\
    ogbl-wikikg2 & 535 & 2,500k & 16,109k & 429k & 598k \\
    ogbl-biokg & 51 & 94k & 4,763k & 163k & 163k \\
    
    \hline
    \end{tabular}
}
\caption{Statistics of Datasets.}
\label{tbl_dataset}
\end{table}

\begin{table*}[hbt!]
\normalsize

\centering
\scalebox{0.76}{
    \begin{tabular}{c|cccc|cccc|cccc}
    \hline
    \textbf{Dataset} & \multicolumn{4}{c|}{\textbf{FB15k-237}} & \multicolumn{4}{c|}{\textbf{DB100K}}  & \multicolumn{4}{c}{\textbf{YAGO37}}\\
    \textbf{Metric} & MRR & HIT@1 & HIT@3 & HIT@10 & MRR & HIT@1 & HIT@3 & HIT@10 & MRR & HIT@1 & HIT@3 & HIT@10 \\
    \hline \hline
    TransE & .294 & - & - & .465 & .111 & .016 & .164 & .270 & .303 & .218 & .336 & .475 \\
    DistMult & .241 & .155 & .263 & .419 & .233 & .115 & .301 & .448 &  .365 & .262 & .411 & .575 \\
    ComplEx & .247 & .158 & .275 & .428 & .242 & .126 & .312 & .440 & .417 & .320 & .471 & .603 \\
    RotatE & .338 & .241 & .375 & .533 & - & - & - & -  & - & - & - & - \\
    SEEK &  .338 & \textbf{.268} & .370 & .467 & .338 & .268 & .370 & .467 & .454 & .370 & .498 & .622 \\
    PairRE & .351 & .256 & .387 & .544 & .412 & .309 & .472 & \textbf{.600} & - & - & - & - \\
    \hline
    
    TranSHER & \textbf{.360} & .264 & \textbf{.397} & \textbf{.551} & \textbf{.431} & \textbf{.345} & \textbf{.476} & .589 &  \textbf{.490} & \textbf{.404} & \textbf{.538} & \textbf{.647} \\
    \hline
    \end{tabular}
}
    \begin{tablenotes}
      \scriptsize
    \item * $p<0.01$ is satisfied in the significance testing.
    \end{tablenotes}

\vspace{-1mm}
\caption{Results on Full Ranking Settings Datasets. Results on FB15k-237 and DB100K are taken from ~\cite{chao_pairre_2021}, while results on YAGO37 are taken from ~\cite{xu-etal-2020-seek}. Dim is referred to the dimension parameter $k$ of entity embeddings.}
\label{tbl_overall_1}
\end{table*}

\noindent \textbf{Baselines.} 
Our baselines include the two main categories of knowledge graph embedding methods for comparison. 
For the similarity-based \textit{semantic matching} methods, DistMult~\cite{yang2014embedding_distmult}, ComplEx~\cite{trouillon2016complex}, and SEEK~\cite{xu-etal-2020-seek} are selected. 
We choose the classic TransE~\cite{bordes_translating_2013_TransE_FB15k}, RotatE~\cite{sun2018rotate}, and the recently proposed PairRE~\cite{chao_pairre_2021} as the baselines in the other category of \textit{distance-based} methods.
The model PairRE is the main baseline in our work.
More details can be referred to in Appendix~\ref{sec:appendix_a_baseline}.

\noindent \textbf{Implementation Details.}
Following our design in \S\ref{sec_method_init}, the three components of TranSHER are initialized with the gamma uniform~\cite{sun2018rotate} and the Xavier normal~\cite{glorot2010understanding_xavier} distribution alternatively to achieve the best result.
Our model is also fine-tuned with light parameter search on $\gamma$ and regularization weights on translation embeddings according to datasets. 
All the experiments are set up in a GPU-accelerated hardware environment.
We follow ~\citet{wang2014TransH} to counts \textit{hpt} and \textit{tph} to categorize relation types through the given dataset $\mathcal{T}$.
Further implementation details could be found in Appendix~\ref{sec:appendix_a_implementaion}.

\begin{table*}[bt]
\normalsize
\centering
\scalebox{0.8}{
\begin{tabular}{c|ccc|ccc}
\hline
\textbf{Dataset} & \multicolumn{3}{c|}{\textbf{ogbl-wikikg2}} & \multicolumn{3}{c}{\textbf{ogbl-biokg}}\\
\textbf{Metric} & Dim & Test MRR & Valid MRR & Dim & Test MRR & Valid MRR \\
\hline \hline
TransE  & 500 & .4256 & .4272  & 2000 & .7452  & .7456 \\
DistMult  & 500 & .3729  & .3506  & 2000 & .8043 & .8055  \\
ComplEx  & 250 & .4027 & .3759  & 1000 & .8095 &  .8105 \\
RotatE & 250 & .4332 & .4353  & 1000 & .7989 & .7997 \\
PairRE & 200 & .5208 & .5423  & 2000 & .8164 & .8172 \\
\hline
TranSHER & 200 & \textbf{.5536} & \textbf{.5662}  & 2000 & \textbf{.8233} & \textbf{.8244} \\

\hline
\end{tabular}
}
\caption{Results on Open Graph Benchmark Link Prediction Datasets. Results of baselines are taken from its official leaderboard~\cite{hu2020_OGB}. The \textit{Dim} column is referred to the dimension parameter $k$ of entity embeddings.}
\vspace{-3mm}
\label{tbl_overall_2}
\end{table*}

\subsection{Overall Results}
As revealed in Tab.~\ref{tbl_overall_1} and Tab.~\ref{tbl_overall_2}, our model achieves significant performance improvement in the main metric MRR on all five different link prediction datasets compared to the strong baselines. 
On the datasets that require full ranking through all entities, TranSHER makes substantial improvement on the main metric MRR as shown in Tab.~\ref{tbl_overall_1}, surpassing PairRE and SEEK. 
As for results on datasets in~\citet{hu2020_OGB}, TranSHER also has considerable performance gain as shown in Tab.~\ref{tbl_overall_2}. 
For the two datasets using the partial ranking setting, TranSHER also achieves incremental performances.
On the very large-scale dataset ogbl-wikikg2, TranSHER improves about 3\% MRR while keeping the dimension number of the entity and thus show its potential to extend on a knowledge graph with a large size. 
On the ogbl-biokg dataset, which distinguishes the challenge by isolating entities by types, TranSHER also shows its superiority at generalizing well across domains.

\subsection{Complex Relations Modeling}\label{sec_result_complex}
The triplets with complex relations (1-to-N, N-to-1, and N-to-N) hold a large portion in many datasets. More importantly, they are more difficult to model than 1-to-1 relations.
In this regard, we conduct experiments to analyze the performance of TranSHER on different types of  triplets.  

\begin{table*}[bt]
\normalsize
\centering
\scalebox{0.75}{
    \begin{tabular}{c|cccc|cccc}
    \hline
    \textbf{Rel. Type} & 1-to-1 & 1-to-N & N-to-1 & N-to-N & 1-to-1 & 1-to-N & N-to-1 & N-to-N \\
    \hline \hline 
    \textbf{Task} & \multicolumn{4}{c|}{\textbf{Head Prediction (MRR)}} & \multicolumn{4}{c}{\textbf{Tail Prediction (MRR)}}  \\
    \hline  
    TransE* & .496 & .457 & .084 & .251 & .482 & .074 & .750 & .365  \\
    DistMult* & .215 & .441 & .074 & .231 & .214 & .052 & .728 & .346 \\
    ComplEx* & .357 & .462 & .092 & .247 & .371 & .060 & .741 & .353 \\
    RotatE* & \textbf{.504} & .467 & .090 & .258 & .488 & .077 & .758 & .373 \\
    PairRE* & .494 & .482 & .112 & .275 & \textbf{.495} & .076 & .766 & .380 \\
    TranSHER & .501 & \textbf{.487} & \textbf{.119} & \textbf{.285} & .494 & \textbf{.079} & \textbf{.779} & \textbf{.389}  \\
    \hline
    \textbf{Task} & \multicolumn{4}{c|}{\textbf{Head Prediction (HIT@10)}} & \multicolumn{4}{c}{\textbf{Tail Prediction (HIT@10)}} \\
    \hline 
    TransE* & .609 & .661 & .166 & .469 & .594 & .138 & .879 & .610 \\
    DistMult* & .453 & .644 & .140 & .422 & .448 & .113 &.844 & .560  \\
    ComplEx* & .521 & .657 & .181 & .453 & .510 & .126  & .861& .588 \\
    RotatE* & .604 & .671 &.170 & .471 & .589 & .146 & .884 & .611  \\
    PairRE* & .599 & .667 & \textbf{.211} & .488 & .589  & .147  & .884 & .619   \\
    TranSHER & \textbf{.615} & \textbf{.674 }& \textbf{.211} & \textbf{.500} & \textbf{.604} & \textbf{.162} & \textbf{.891} & \textbf{.624}  \\
    \hline
    
    \end{tabular}
}
\caption{Evaluation Result on Different Relation Types on FB15k-237. Models with $*$ are reproduced with the self-adversarial framework~\cite{sun2018rotate}.}

\label{tbl_rel_type_result}
\end{table*}
\begin{figure*}[bt]



\centering
\begin{subfigure}{\linewidth}
\centering
\includegraphics[width=0.65\linewidth]{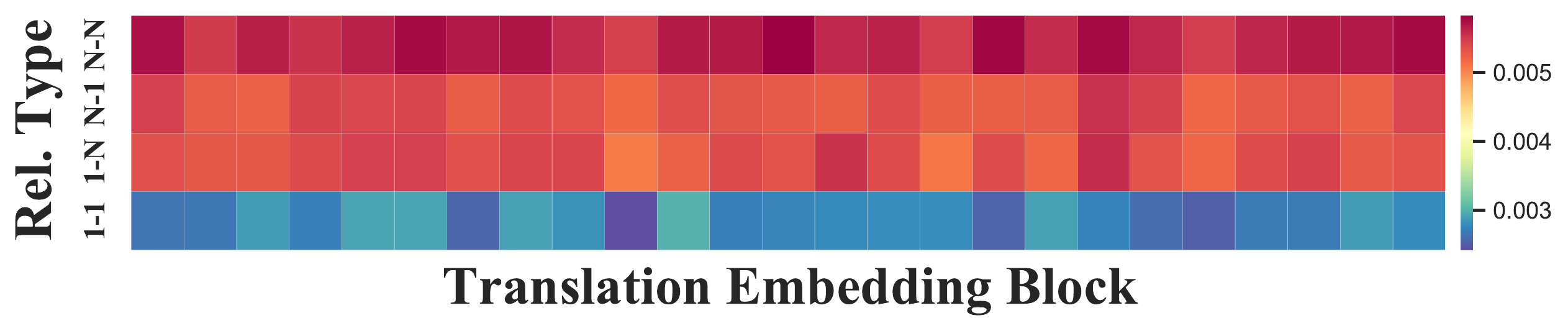}
\caption{FB15K-237}\label{bias_embedding_meanpooling_abs:a}
\end{subfigure}
\begin{subfigure}{\linewidth}
\centering
\includegraphics[width=0.65\linewidth]{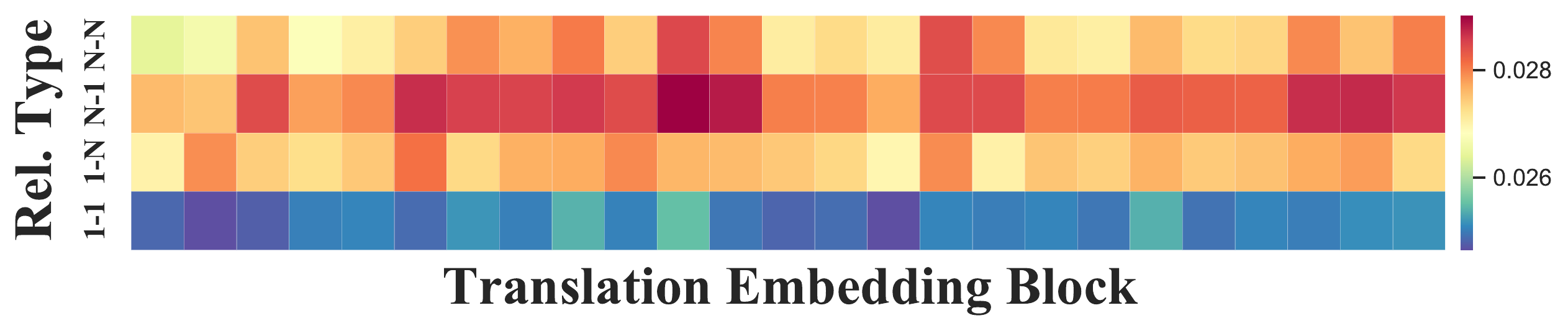}
\caption{DB100K}\label{bias_embedding_meanpooling_abs:b}
\end{subfigure}

\caption{Heat Map Of Translation Embedding Values. The translation embeddings are first grouped by their relation types and their average absolute values are calculated along each dimension. We then implement a mean pooling operation on relation-type-wide average bias embeddings. Specifically, the pooling block size is set to 60 for the FB15k-237 model (1500-dim) and 20 for the DB100K model (500-dim).}
\label{bias_embedding_meanpooling_abs}
\end{figure*}

Results on FB15k-237 along relation types in Tab. \ref{tbl_rel_type_result} demonstrate that TranSHER makes stable gains on triplets with complex relations. For the N-to-N relation triplets (accounting for 87\% of the whole dataset), TranSHER achieves better performances in MRR and HIT@10 on both head prediction and tail prediction tasks, even compared to the best baseline PairRE. This signifies that our proposed model can actually model complex relations better. A potential reason for this improvement of TranSHER is its most distinct part, relation-specific translation. We will give a more detailed analysis in the following sections.

\vspace{-1mm}
To further learn the intrinsic influence of the relation-specific translation of TranSHER on different relation types, we visualize the translation embeddings on the FB15k-237 and DB100K by 
presenting the absolute value heat maps.
As shown in Fig.~\ref{bias_embedding_meanpooling_abs}, the translation embeddings of complex relations have obvious color differences from those 1-to-1 relation embeddings. The embeddings of 1-to-1 relations are closer to zero than complex relations; this signifies the translational item mainly contributes to complex relations 1-to-N, N-to-1, and N-to-N, whilst 1-to-1 relations are learned less. 
Specifically, translations of N-to-N relations are the most active, which suggests TranSHER puts more effort into these hard-to-learn relations. The behavior of translations on 1-to-1 relations implies that the easier relations require less optimization during training. 
The observation of this preference for complex relations from the relation-specific translations provides supportive evidence for the experimental results in Tab. ~\ref{tbl_rel_type_result}. 

\begin{figure}[bt]

\centering

\begin{subfigure}{0.9\linewidth}
\includegraphics[width=\linewidth]{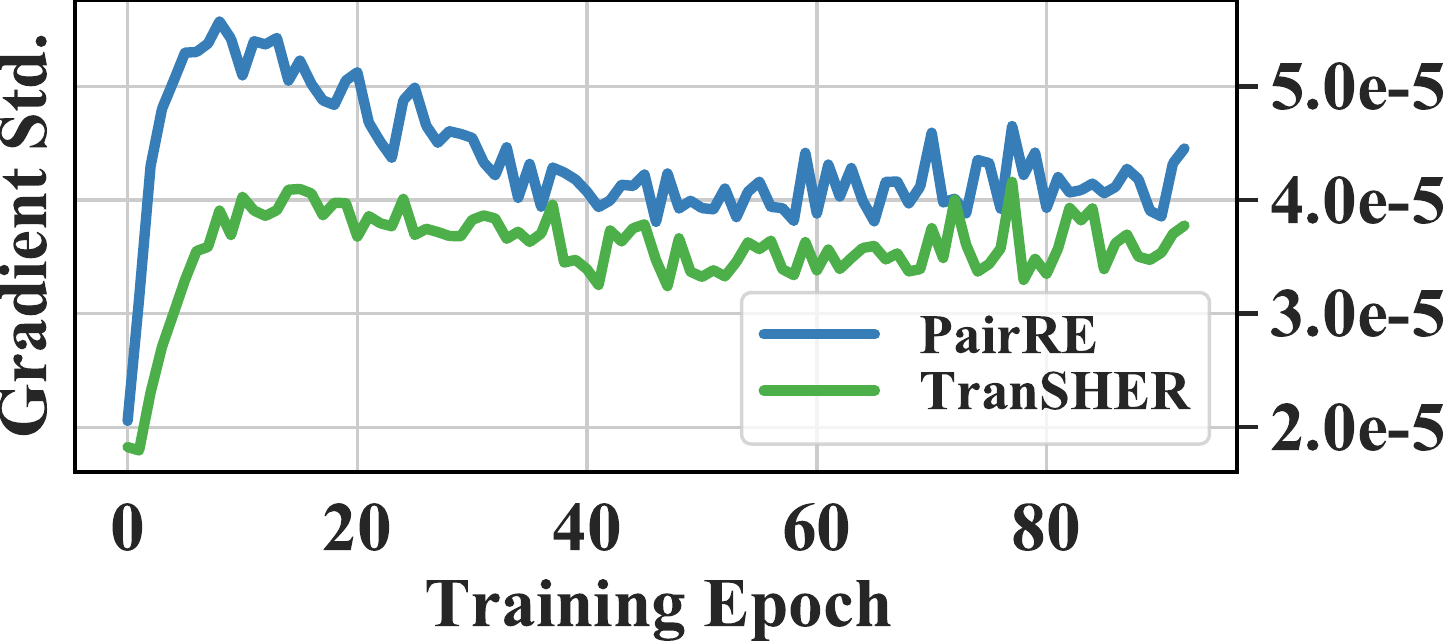} 
\caption{Entity}\label{gradient_plot:a}
\end{subfigure}
\begin{subfigure}{0.9\linewidth}
\includegraphics[width=\linewidth]{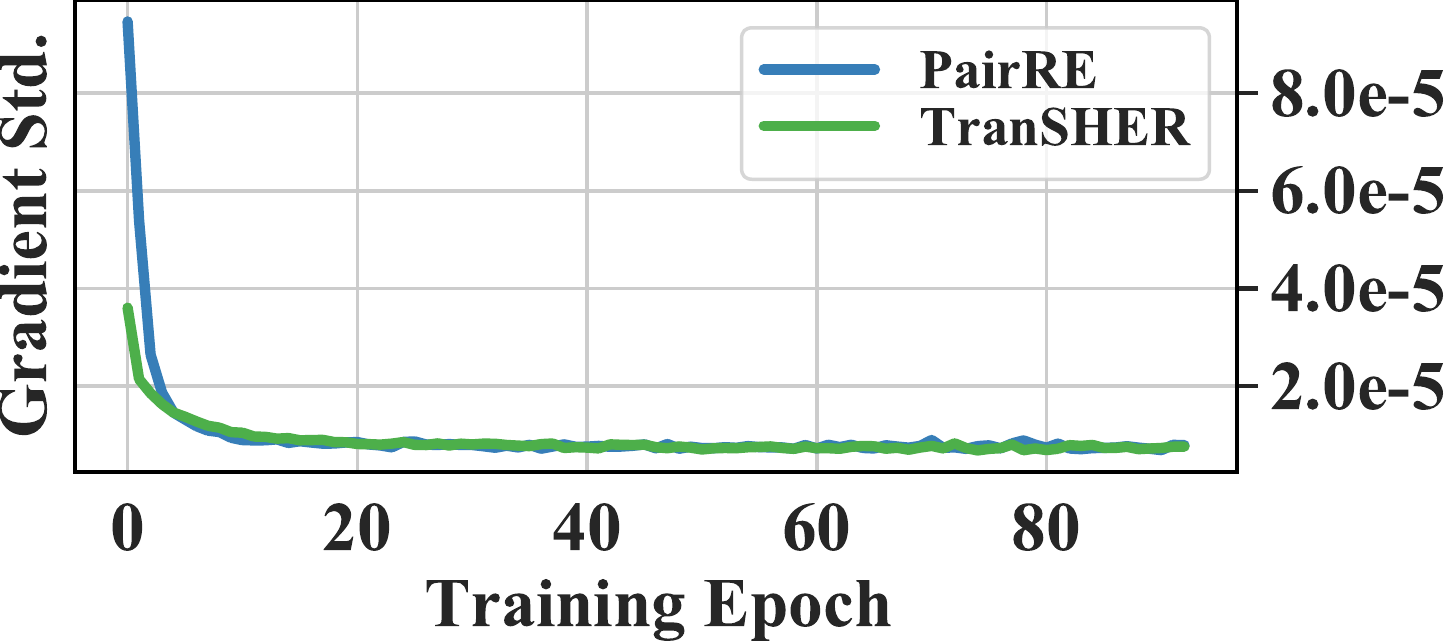} 
\caption{Relation}\label{gradient_plot:b}
\end{subfigure}

\caption{
Statistics of Model Gradients in FB15k-237 Training. We plot the standard deviation of weight gradients for entity and relation embeddings at the beginning of each training epoch. Gradient standard deviation of translations is not calculated and plotted since the baseline model does not contain such a module.
}
\label{gradient_plot}
\end{figure}
\begin{figure}[tb]
\centering
\begin{subfigure}{0.48\linewidth}
\centering
\includegraphics[width=\linewidth]{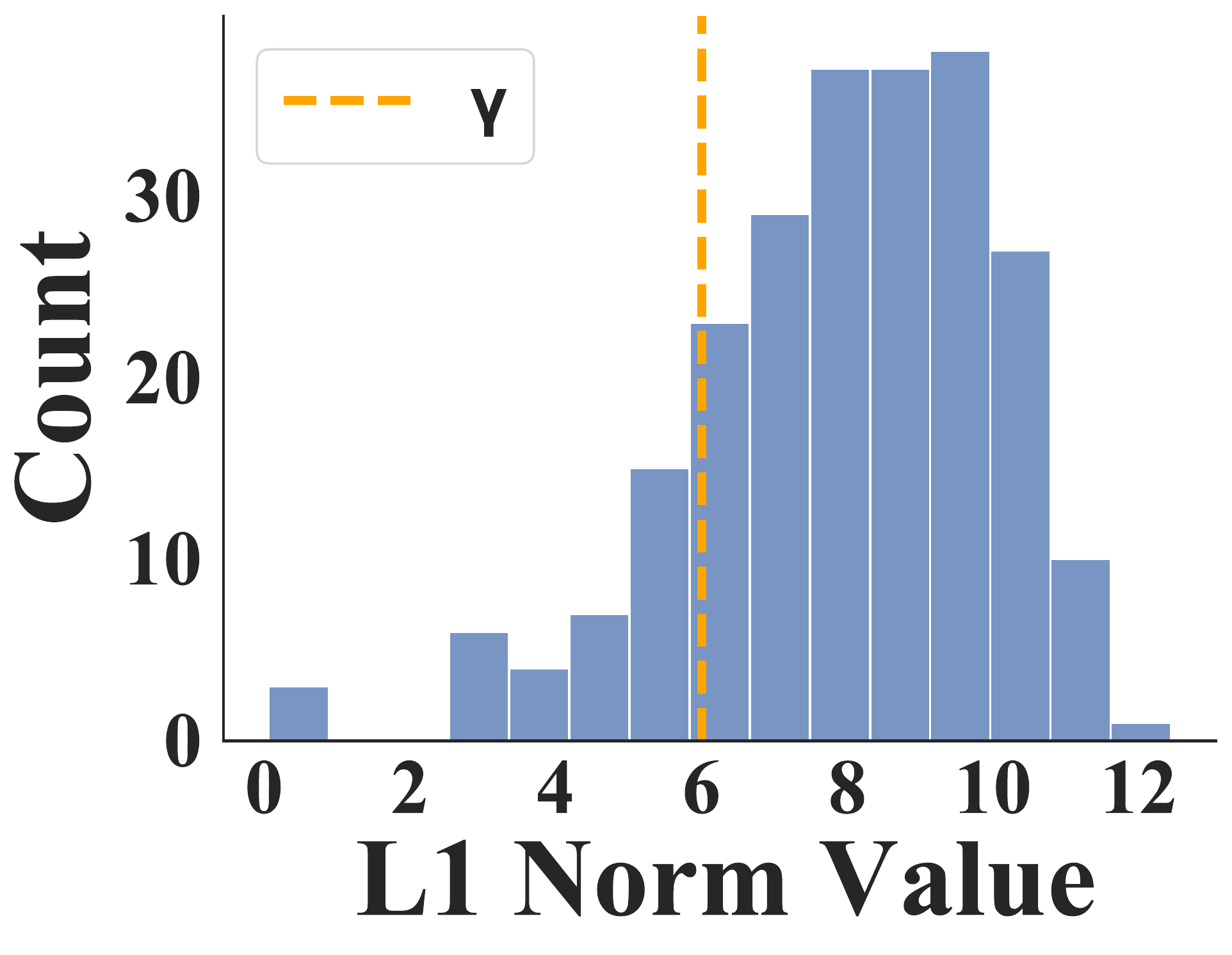} 
\caption{FB15K-237}\label{bias_L1:a}
\end{subfigure}
\begin{subfigure}{0.48\linewidth}
\centering
\includegraphics[width=\linewidth]{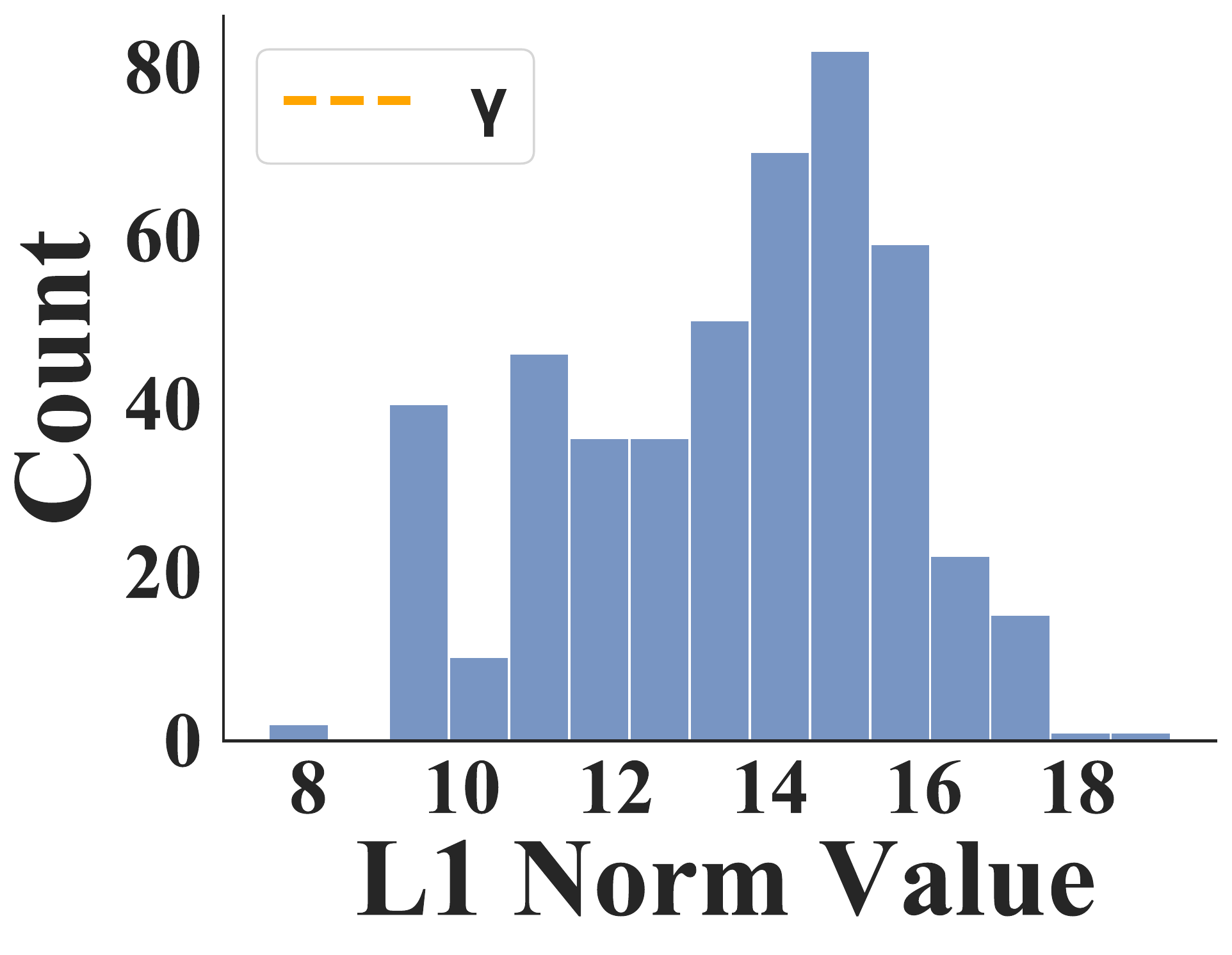} 
\caption{DB100K}\label{bias_L1:b}
\end{subfigure}
\caption{Histogram of the L1 Norm of TranSHER Translations on two dataset.}

\label{bias_L1}
\end{figure}
\begin{table*}[hbt!]
\normalsize

\centering
\scalebox{0.8}{
    \begin{tabular}{ccc|cccc| cccc}
    \hline
    \multicolumn{3}{c|}{\textbf{Initialization}}  &  \multicolumn{4}{c|}{\textbf{FB15k-237}}&  \multicolumn{4}{c}{\textbf{DB100K}} \\ 
    \hline
    $\mathcal{E}$ & $G_r$ & $B_r$ & MRR & HIT@1 & HIT@3 & HIT@10  & MRR & HIT@1 & HIT@3 & HIT@10 \\
    \hline \hline
    
    $\mathcal{N_{X}}$  &  $\mathcal{N_{X}}$  & $\mathcal{N_{X}}$  & .357 & .262 & .394 & .548 & .426 & .336 & .475 & \textbf{.592} \\  
    \hdashline 
    $\mathcal{N_{X}}$  &  $\mathcal{N_{X}}$  & $\mathcal{U}_{\gamma}$ & .353 & .260 & .389 & .542 & .429 & .345 & .474 & .585 \\  
    $\mathcal{N_{X}}$  &  $\mathcal{U}_{\gamma}$ & $\mathcal{N_{X}}$  & .353 & .258 & .389 & .545 & \textbf{.431} & .345 & \textbf{.476} & .589 \\  
    $\mathcal{U}_{\gamma}$ & $\mathcal{N_{X}}$  &  $\mathcal{N_{X}}$  &  \textbf{.360} & \textbf{.264} & \textbf{.397} & \textbf{.551} & .423 & .335 & .470 & .585  \\  
    \hdashline 
    $\mathcal{N_{X}}$  &  $\mathcal{U}_{\gamma}$ & $\mathcal{U}_{\gamma}$ &  .347 & .255 & .381 & .533  & .430 & \textbf{.347} & .473 &. 583\\  
    $\mathcal{U}_{\gamma}$ & $\mathcal{N_{X}}$  & $\mathcal{U}_{\gamma}$ &  .355 & .261 & .389 & .543 & .425 & .343 & .470 & .580 \\  
    $\mathcal{U}_{\gamma}$ &  $\mathcal{U}_{\gamma}$ & $\mathcal{N}_{X}$ & .357 & .262 & .394 & .548 & .424 & .338 & .469 & .583 \\
    \hdashline 
    $\mathcal{U}_{\gamma}$ & $\mathcal{U}_{\gamma}$ & $\mathcal{U}_{\gamma}$ &  .348 & .255 & .384 & .535 & .424 & .341 & .468 & .577 \\  
    \hline
    \end{tabular}
}
\caption{A Study of Different Initialization Strategies for TranSHER. The experiments are conducted on FB15k-237 and DB100K datasets with full ranking settings. The results are grouped by the number of gamma uniform or Xavier normal distributions used in the combinations.}
\label{component_initialization}
\end{table*}

\subsection{Translation Impacts}
In order to explore how exactly the relation-specific translation affects modeling, we analyze the influence of the translations from the perspectives of training optimization and the score function itself.

To study TranSHER from the view of the training process, the standard deviation of weight gradients for entities and relations are plotted in Fig.~\ref{gradient_plot}, following similar settings in~\citet{glorot2010understanding_xavier}. We found that TranSHER largely reduces the gradient standard deviation of relation embeddings only at the beginning of training and keeps a similar trend to the baseline for the rest epochs. What most distinguishes the optimization process of TranSHER from PairRE is that TranSHER maintains a relatively low standard deviation of entity embedding gradients along with the whole training. Such a low standard deviation implies the stable optimization progress of TranSHER. 
The adjustment of entity embeddings usually requires more effort since the number of entities is about thirty times larger than the relations in FB15k-237 (and similar cases for other datasets, see Tab.~\ref{tbl_dataset}). We suspect that the generally superior results of TranSHER are mainly brought about by a more stable optimization of entity embeddings.

From the perspective of the score function, the relation-specific translation item can directly affect the distance calculation. We plot the distribution of L1 Norm values of the $k$-dimensional translation embeddings to learn of such an impact (Fig.~\ref{bias_L1}). Most L1 Norms of translation embeddings are larger than the margin $\gamma$ in both experiments on FB15k-237 and DB100K, which suggests a prominent impact from the translation in modeling.

\subsection{Initialization Strategy}\label{sec_result_bias_init}

During the implementation, we found that the initial strategy of TranSHER components is crucial to the performance, and thus we conduct a further study on the initialization strategies combinations on the FB15k-237 and DB100K datasets. 

We compare the results of several initialization methods to learn the effect of different distributions, as revealed in Tab.~\ref{component_initialization}.
Specifically, the gamma uniform distribution $\mathcal{U}_{\gamma}$ and the Xavier normal distribution $\mathcal{N_{X}}$ are alternatively adopted for three components in TranSHER, i.e. the entities $\mathcal{E}$, the relational mapping $G_r$, and the relation-specific translations $B_r$. Such a strategy of initialization combination produces a set of eight experiments on each dataset while keeping the same size of model parameters.

Among all the initialization combinations, the strategy of '{\small$ \mathcal{U}_{\gamma}\mathcal{N_{X}}\mathcal{N_{X}}$}' leads to the strongest performance on FB15K-237, while the '{\small$ \mathcal{N_{X}}\mathcal{U}_{\gamma}\mathcal{N_{X}}$}' variant gets the best MRR result on DB100K. 
The observation of the best initialization strategy varying through datasets suggests that the initialization strategy of TranSHER can accommodate discrepancies through different knowledge graphs, which brings about performance gains on link prediction.

\begin{table}[bt]
\centering
\scalebox{0.68}{
\begin{threeparttable}\normalsize
    \begin{tabular}{c|ccccc}
    \hline %
    \textbf{Query} & \multicolumn{5}{|c}{(Cinderella,  /film/film/produced\_by,  \textbf{?})} \\
    \hline
    \textbf{Answer} & \multicolumn{5}{|c}{Walt Disney} \\
    \hline 
    \textbf{Model} &  \multicolumn{2}{c|}{TranSHER} &  \multicolumn{3}{c}{PairRE} \\
    \hline \hline
    Rank 1 & \multicolumn{2}{c|}{\textcolor{teal}{$\bullet$ \textbf{Walt Disney}}}  & \multicolumn{3}{c}{Walt Disney Animation Studios} \\
    Rank 2 & \multicolumn{2}{c|}{Ivan Reitman}  & \multicolumn{3}{c}{The Walt Disney Company} \\
    Rank 3 &  \multicolumn{2}{c|}{Hayao Miyazaki} & \multicolumn{3}{c}{Jerry Bruckheimer} \\
    Rank 4 & \multicolumn{2}{c|}{Jerry Bruckheimer}  & \multicolumn{3}{c}{Alan Menken} \\
    Rank 5 & \multicolumn{2}{c|}{Walt Disney Pictures}  & \multicolumn{3}{c}{\textcolor{teal}{$\bullet$\textbf{Walt Disney}}} \\
    Rank 6 & \multicolumn{2}{c|}{Gary Goetzman}  & \multicolumn{3}{c}{Hayao Miyazaki} \\
    Rank 7 & \multicolumn{2}{c|}{Lawrence Golden}  & \multicolumn{3}{c}{Walt Disney Picture} \\
    Rank 8 & \multicolumn{2}{c|}{Howard Ashman}  & \multicolumn{3}{c}{John Lasseter} \\
    \hline
    \end{tabular}
\begin{tablenotes}
\small
  \item * The $\bullet$ refers to the correct answer.
\end{tablenotes}
\end{threeparttable}
}
\caption{A Case Study of Tail Prediction on FB15k-237.}\label{tbl_case}
\end{table}

\subsection{Case Study}\label{sec_case}
We further provide a case study to illustrate the effectiveness of TranSHER in handling challenging link predictions. 
In Tab. \ref{tbl_case}, the query asks for the producer of a 1950 animated musical fantasy film \textit{Cinderella} (NB: the produced\_by relation defined in FB15k-237  \textit{only} refers to producers). 
While PairRE is capable of retrieving relevant entities like the production studio/company, and even a composer that used to work on another \textit{Walt Disney} film, it still struggles with learning the semantic meaning of the relation and mixing the neighbor entities in representation space.
Meanwhile, the high-ranking entities found by TranSHER are mostly producers/directors and the exact subsidiary of the production studio. 
This implies that TranSHER does not only have the ability to cluster the relevant entities restricted on hyper-ellipsoids but can also accurately model the semantic meaning of the particular relation "who is the producer of the film" with the relation-specific translation item.
More cases can be found in Appendix~\ref{sec:appendix_b}.

\section{Conclusion}
We propose a novel knowledge graph embedding model TranSHER for the link prediction task. 
TranSHER leverages relation-specific translation on entities with hyper-ellipsoidal restriction, which is explicitly encoded into the score function. %
By introducing the translation, TranSHER can improve the optimization of entities distributed on hyper-ellipsoids and shows ingenuity in understanding semantic characteristics. 
Moreover, we prove that TranSHER preserves the ability to represent logical reasoning relation patterns.
Comprehensive experiments on different datasets show that TranSHER has robust performance and improves complex relation modeling. 

\section*{Limitations}
The proposed model TranSHER mainly provides insight into how the translation item can improve the knowledge graph embedding methods for the link prediction task with restricted entities. However, due to the enormous number of entities in knowledge graphs, this work does not directly show the learned entity representation distribution, which could potentially provide further information beneficial to the task.
\section*{Acknowledgement}
Yizhi Li is fully funded by an industrial PhD studentship (Grant number: 171362) from the University of Sheffield, UK.

\bibliographystyle{acl_natbib}
\bibliography{reference}

\appendix

\clearpage
\section{Experimental Details} \label{sec:appendix_a}
\subsection{Datasets}\label{sec:appendix_a_data}


Extensive experiments are conducted on five publicly available datasets. 
The FB15k-237 dataset is filtered from another dataset FB15k~\cite{toutanova_observed_2015_fb15k237, bordes_translating_2013_TransE_FB15k}, which is built from a knowledge fact database, Freebase~\cite{bollacker2008freebase}. 
Compared to FB15k-237, two larger-scale knowledge graphs DB100K~\cite{ding-etal-2018-improving-db100k} and YAGO37~\cite{guo2018knowledge_RUGE_YAGO37} with about ten times of entities are also selected. 
The YAGO37 is a subset selected from the YAGO3 core facts~\cite{guo2018knowledge_RUGE_YAGO37,suchanek2007yago}. 
and the DB100K is constructed from the mapping-based
objects of core DBpedia~\cite{ding-etal-2018-improving-db100k, bizer2009dbpedia}.
We additionally test our model on two distinguished link prediction datasets~\cite{hu2020_OGB}.
ogbl-wikikg2 is a very large-scale dataset derived from the Wikidata knowledge base~\cite{vrandevcic2014wikidata}. ogbl-biokg uses data from biomedical data repositories and divides the entities into 5 types according to domain knowledge.

To prove the generalization ability of TranSHER, we select datasets of various entity quantities from 15k (FB15k-237) to 2,500k (ogbl-wikikg2), which spans three orders of magnitude. 

\subsection{Evaluation Protocol}\label{sec:appendix_a_eval}
Following standard implementation, we use Mean Reciprocal Rank (MRR) and Hits at N (HIT@N) as our metrics on all the datasets, whilst MRR is referred to as the main evaluation measure due to its relatively comprehensive perspective. The general evaluation settings for the link prediction task can be recognized as the full ranking setting and the partial ranking setting according to the selection methods of negative samples for testing.

Except for datasets selected from~\citet{hu2020_OGB}, the ranking candidates are all entities that appeared in the knowledge graph, i.e. a full ranking task setting. 
Following the standard protocol, the scores of the extra correct entities in each query are filtered during full ranking testing.
Note that for ogbl-wikikg2 and ogbl-biokg, the ranking candidates are the positive entity and 500 randomly sampled negative entities (with a separate 500 for prediction head and tail tasks). Specifically, the sampled negative entities belong to the same type of positive ones in the ogbl-biokg.

\subsection{Baselines}\label{sec:appendix_a_baseline}
Two main categories of knowledge graph embedding methods are chosen in our work to compare and validate the performance of TranSHER:

\begin{itemize}[noitemsep,topsep=5pt]
    \item Semantic matching score functions
    \item Distance-based score functions
\end{itemize}

The semantic matching score functions intend to study interactions in knowledge graphs with similarity-based score functions. We select DistMult~\cite{yang2014embedding_distmult} and ComplEx~\cite{trouillon2016complex} as the representatives of such methods that leverage inner-product largely. Moreover, a recently proposed framework SEEK~\cite{xu-etal-2020-seek} is selected as a strong baseline for the dataset YAGO37. We also categorize SEEK as a semantic matching due to its usage of inner-product calculation even if it conducts hard segmentation on the embeddings.

Methods in the other category, \textit{distanced-based}, design score functions to model the distance between connected entities in low-dimensional space. Additional to the foundational distance-based function TransE~\cite{bordes_translating_2013_TransE_FB15k}, we also take two more effective models RotatE~\cite{sun2018rotate} and the PairRE~\cite{chao_pairre_2021} as our baselines. 
PairRE is the main baseline since it has competitive performance and adopts the same restriction on the entities as TranSHER. 


\subsection{Implementation Details}\label{sec:appendix_a_implementaion}
As described in \S\ref{sec_method_init}, the entity embeddings in $\mathcal{E}$, mapping weights of $G_r$, and relation-specific translations $B_r$ can be initialized either with gamma uniform implemented in ~\cite{sun2018rotate, chao_pairre_2021} or the Xavier normal distribution~\cite{glorot2010understanding_xavier}.
The gamma uniform is set as $ \mathcal{U}_{\gamma}(-\frac{\gamma+\epsilon}{k}, \frac{\gamma+\epsilon}{k}), \epsilon=2.0$ the scaling gain for Xavier normal $\mathcal{N}_{X}(0,g \cdot \sqrt{\frac{2}{k}})$ is set to $g=1.0$. 
Parameter searches on $\gamma$ and regularization weights on translation embeddings are adopted, while the embedding dimension $k$ remains the same as PairRE. The extra effort of embedding dimension tuning is spent on the YAGO37 dataset since the original work PairRE does not provide the implementation on it.

All the experiments occupy a capacity under 16GB RAM on an RTX 3090 GPU. 
To get the relation type information following ~\citet{wang2014TransH}, we count \textit{hpt} and \textit{tph} through all the triplets from the given dataset including the test split. 

\begin{table*}[htbp]
\centering
\begin{threeparttable}\small
    \begin{tabular}{c|ccccc|ccccc}
    \hline %
    \textbf{Query} & \multicolumn{5}{|c|}{(\textbf{?}, /medicine/symptom/symptom\_of, jaundice)}& \multicolumn{5}{|c}{(\textbf{?}, /music/group\_membership/group, USA for Africa)} \\
    \hline
    \textbf{Answer} & \multicolumn{5}{|c}{vomiting} & \multicolumn{5}{|c}{Stevie Wonder} \\
    \hline 
    \textbf{Model} &  \multicolumn{2}{c|}{TranSHER} &  \multicolumn{3}{c}{PairRE}  &  \multicolumn{2}{|c|}{TranSHER} &  \multicolumn{3}{c}{PairRE} \\
    \hline \hline
    Rank 1 & \multicolumn{2}{c|}{\textcolor{teal}{$\bullet$ \textbf{vomiting}}}  &   \multicolumn{3}{c|
    }{pancreatic cancer} & \multicolumn{2}{|c|}{Janet Jackson }  & \multicolumn{3}{c}{USA for Africa}  \\
    Rank 2 & \multicolumn{2}{c|}{dyspnea}  & \multicolumn{3}{c}{liver tumor} 
    & \multicolumn{2}{|c|}{Norah Jones }  & \multicolumn{3}{c}{Michael Sembello}  \\
    Rank 3 &  \multicolumn{2}{c|}{fever} & \multicolumn{3}{c}{ liver cirrhosis} 
    & \multicolumn{2}{|c|}{\textcolor{teal}{$\bullet$ \textbf{Stevie Wonder}}} & \multicolumn{3}{c}{Elton John}  \\ 
    Rank 4 & \multicolumn{2}{c|}{liver cirrhosis}  & \multicolumn{3}{c}{ fever} 
    & \multicolumn{2}{|c|}{Quincy Jones }  & \multicolumn{3}{c}{ Bobby Darin} \\
    Rank 5 & \multicolumn{2}{c|}{diarrhea }  & \multicolumn{3}{c}{hepatitis B} & \multicolumn{2}{|c|}{Prince}  & \multicolumn{3}{c}{Norah Jones}  \\
    Rank 6 & \multicolumn{2}{c|}{anorexia}  & \multicolumn{3}{c}{diarrhea} 
    & \multicolumn{2}{|c|}{Michael McDonald }  & \multicolumn{3}{c}{Irene Cara}\\ 
    Rank 7 & \multicolumn{2}{c|}{headache}  & \multicolumn{3}{c}{\textcolor{teal}{$\bullet$ \textbf{vomiting}}} 
    & \multicolumn{2}{|c|}{USA for Africa}  & \multicolumn{3}{c}{\textcolor{teal}{$\bullet$ \textbf{Stevie Wonder}}}\\  
    Rank 8 & \multicolumn{2}{c|}{pancreatic cancer}  & \multicolumn{3}{c}{headache}  
    & \multicolumn{2}{|c|}{Barbra Streisand}  & \multicolumn{3}{c}{Quincy Jones}\\ 
    \hline
    \end{tabular}
\begin{tablenotes}
\scriptsize
  \item * The $\bullet$ refers to the correct answer.
\end{tablenotes}
\caption{Additional Cases for Head Prediction Task on FB15k-237.}\label{tbl_case_more_head}
\end{threeparttable}
\end{table*}

\begin{table*}[htbp]
\centering
\begin{threeparttable}\small
    \begin{tabular}{c|ccccc|ccccc}
    \hline 
    \textbf{Query} & \multicolumn{5}{|c|}{( Dena Higley,  /people/person/profession, \textbf{?})}& \multicolumn{5}{|c}{(Almost Famous, /film/film/genre, \textbf{?})} \\
    \hline
    \textbf{Answer} & \multicolumn{5}{|c}{writer} & \multicolumn{5}{|c}{ coming of age} \\
    \hline 
    \textbf{Model} &  \multicolumn{2}{c|}{TransHER} &  \multicolumn{3}{c}{PairRE}  &  \multicolumn{2}{|c|}{TransHER} &  \multicolumn{3}{c}{PairRE} \\
    \hline \hline
    Rank 1 & \multicolumn{2}{c|}{\textcolor{teal}{$\bullet$ \textbf{writer}}}  & \multicolumn{3}{c|
    }{television producer} & \multicolumn{2}{|c|}{romance film }  & \multicolumn{3}{c}{romance film}  \\
    Rank 2 & \multicolumn{2}{c|}{television producer}  & \multicolumn{3}{c}{actor} & \multicolumn{2}{|c|}{\textcolor{teal}{$\bullet$ \textbf{coming of age}}}  & \multicolumn{3}{c}{LGBT}  \\
    Rank 3 &  \multicolumn{2}{c|}{actor} & \multicolumn{3}{c}{ television director} & \multicolumn{2}{|c|}{independent film} & \multicolumn{3}{c}{ historical period drama}  \\ 
    Rank 4 & \multicolumn{2}{c|}{journalist}  & \multicolumn{3}{c}{film director} & \multicolumn{2}{|c|}{romantic comedy}  & \multicolumn{3}{c}{ independent film} \\
    Rank 5 & \multicolumn{2}{c|}{film director}  & \multicolumn{3}{c}{film producer} & \multicolumn{2}{|c|}{black comedy}  & \multicolumn{3}{c}{romantic comedy}  \\
    Rank 6 & \multicolumn{2}{c|}{television director}  & \multicolumn{3}{c}{television presenter} & \multicolumn{2}{|c|}{ LGBT}  & \multicolumn{3}{c}{biography}\\ 
    Rank 7 & \multicolumn{2}{c|}{author}  & \multicolumn{3}{c}{\textcolor{teal}{$\bullet$ \textbf{writer}}} & \multicolumn{2}{|c|}{ historical period drama}  & \multicolumn{3}{c}{biographical film}\\  
    Rank 8 & \multicolumn{2}{c|}{film producer}  & \multicolumn{3}{c}{journalist}  & \multicolumn{2}{|c|}{ biographical film }  & \multicolumn{3}{c}{\textcolor{teal}{$\bullet$ \textbf{coming of age}}}\\ 
    \hline
    \end{tabular}
\begin{tablenotes}[flushleft]\footnotesize
\scriptsize 
  \item [*] The $\bullet$ refers to the correct answer.
\end{tablenotes}
\caption{Additional Cases for Tail Prediction Task on FB15k-237.}\label{tbl_case_more_tail}
\end{threeparttable}

\end{table*}

\section{Supplementary Case Study}\label{sec:appendix_b}
We provide extra cases for further analysis on the FB15k-237 dataset, where the head prediction task cases are shown in Tab.~\ref{tbl_case_more_head} and tail prediction tasks in Tab.~\ref{tbl_case_more_tail}, respectively.

In general, we can observe that both TranSHER and PairRE can retrieve relevant entities in the same or similar topic according to the given relation-entity queries, which distribute from the entertainment domain to medical knowledge facts. For instance, in the \textit{jaundice} case at the left of Tab.~\ref{tbl_case_more_head}, the entities recalled by the models are all terminologies from the practice of medicine.
This implies that the score functions with hyper-ellipsoidal restriction can model the entities in the same neighborhood and assign close positions in the latent space with regard to their graphical interactions such as the n-hop distances.

However, similar to the case described in \S\ref{sec_case}, TranSHER shows additional precision for predicting the correct entities by learning the inherent semantic categorization of entities with the extra relation-specific translations. Regarding the aforementioned \textit{jaundice} example, the query specifically asks for the symptom of the jaundice disease, which could be highly related to liver diseases.\footnote{{\scriptsize More information can be found at the \href{https://en.wikipedia.org/wiki/Jaundice}{Jaundice wikipage}.}}
Although the PairRE has learned the close relationships between jaundice and liver diseases and distributed these entities close with the hyper-ellipsoidal mapping, it fails to distinguish the concept disease from the close concept symptom and assigns the liver-related with high ranks.
In contrast with PairRE, our proposed model TranSHER can learn the nuanced differences in these two concepts defined in the relations. 
As a result, we suspect that the semantic characteristic modeling of score functions can be improved by the additional relation-specific translation item in TranSHER.

\end{document}